\documentclass{article}
\usepackage{enumitem}
\usepackage{arxiv}

\usepackage[utf8]{inputenc} 
\usepackage[T1]{fontenc}    
\usepackage{hyperref}       
\usepackage{booktabs}       
\usepackage{nicefrac}       
\usepackage{microtype}      
\usepackage{lipsum}
\usepackage{multirow}
\usepackage[table]{xcolor}
\usepackage{siunitx}
\usepackage{makecell} 
\usepackage{svg}
\usepackage{algpseudocode}
\usepackage{algorithm}
\usepackage{array}
\usepackage[caption=false,font=normalsize,labelfont=sf,textfont=sf]{subfig}
\usepackage{textcomp}
\usepackage{arydshln}
\usepackage{stfloats}
\usepackage{url}
\usepackage{verbatim}
\usepackage{rotating}
\usepackage{graphicx}
\usepackage{multicol}
\usepackage{cellspace}
\usepackage{subcaption}
\usepackage{amsmath}        
\usepackage{amssymb}        
\usepackage{algorithm}      
\usepackage{algorithmicx}   
\usepackage{algpseudocode}  
\usepackage{algpseudocode}
\usepackage{algorithm}
\usepackage{array}
\usepackage[caption=false, ,font=normalsize, ,labelfont=sf, ,textfont=sf]{subfig}
\usepackage{textcomp}
\usepackage{stfloats}
\usepackage{url}
\usepackage{verbatim}
\usepackage{multirow}
\usepackage{rotating}
\usepackage{graphicx}
\usepackage{multicol}
\usepackage{multirow}
\usepackage{booktabs}
\usepackage{cellspace}
\usepackage{subcaption}
\usepackage{wrapfig}

\usepackage{amsmath, amsfonts, amsthm, amssymb}

\title{Important Equations}
\author{itxwaleedrazzaq }

\title{\Large Developing Distance-Aware Physics-Constrained Probabilistic Frameworks for Industrial Prognostics}

\author{
  Waleed Razzaq  \\
  School of Automation\\
  University of Science and Technology China\\
  Hefei, Anhui \\
  \texttt{waleed.razzaq@mail.ustc.edu.cn} \\
   \And
 Yun-Bo Zhao \thanks{Corresponding author. Email: \texttt{ybzhao@ustc.edu.cn}} \\
  School of Automation\\
  University of Science and Technology China\\
  Hefei, Anhui \\
  \texttt{ybzhao@ustc.edu.cn} \\
}

\begin{document}
\maketitle
\begin{abstract}
Development of reliable and physically interpretable probabilistic frameworks for industrial prognostics remain nascent, and existing literature is often insensitive as inputs move away from the training manifold. In this paper, we develop two sampling-free, distance-aware physics-constrained probabilistic frameworks: (i) PC-SNGP and (ii) PC-SNER. Both apply spectral normalization to hidden layer weights, enforcing bi-Lipschitz distance-preserving representation from the input to the latent space. PC-SNGP replaces the dense output with Gaussian process whose posterior variance increases with input distance from the training manifold. PC-SNER modifies the output layer to predict Normal-Inverse-Gamma~(NIG) parameters for distance preserving estimation. To maintain balance between data fidelity and physical consistency during training, we introduce a dynamic weighting strategy for the physics-constrained loss. We also introduce a distance-aware-coefficient~(DAC) metric to quantify sensitivity to distributional shifts. Empirically, we validate both frameworks on rolling-element-bearings (REBs) prognostics using the PRONOSTIA, XJTU-SY, and HUST benchmark datasets. Experimental results demonstrate improved prediction accuracy and well-calibrated uncertainty estimates relative to competing baselines, while maintaining auditable performance in cross-validation and robustness under extreme adversarial perturbations.
\end{abstract}


\section{Introduction}

Uncertainty quantification (UQ) in industrial prognostics aims to characterize the confidence of model predictions. Predictive uncertainty is commonly decomposed into two sources: (i) \textit{aleatoric}, arising from inherent, irreducible variability in measurements; and (ii) \textit{epistemic}, arising from limited data or incomplete knowledge~\cite{gawlikowski2023survey}. Accurate characterization of both uncertainties is as critical as prediction accuracy, since overconfident predictions can lead to unsafe maintenance decisions, causing significant economic losses or safety risks. Rotating machinery is particularly prone to failure due to rolling-element bearings (REBs), which are estimated to account for 40--50\% of failures~\cite{ding2021remaining} because of their operation under severe thermal and mechanical stresses. Therefore, it is necessary to develop a prognostic systems that can effectively monitor bearing degradation while reliably quantifying predictive uncertainty.

Existing literature can be broadly classified into three categories: (i) physics-based models (PBMs), (ii) data-driven models (DDMs), and (iii) physics-constrained models (PCMs). PBMs ensure consistency with fundamental physical principles but are limited in representing uncertainty and often lack robustness when applied to complex systems. DDMs typically produce deterministic predictions~\cite{guo2022remaining, liu2023prediction, yang2024rolling, qi2024anomaly, razzaq2025carle}; even probabilistic extensions~\cite{rivas2022predictions, song2024remaining, ochella2024bayesian} primarily capture the output mean and variance and exhibit limited sensitivity to distributional shifts. PCMs incorporate known physical laws into the learning process~\cite{raissi2019physics}, improving generalization and interpretability relative to both PBMs and DDMs. However, existing probabilistic PCM approaches, including Bayesian methods~\cite{linka2022bayesian}, Monte Carlo dropout~\cite{alhajeri2022physics}, and Deep-Ensembles~\cite{soibam2024inverse} lack distance awareness. As a result, they do not reliably increase predictive uncertainty as inputs move away from the training manifold, limiting their effectiveness under OOD conditions.

To advance this, we develop two sampling-free, distance-aware physics-constrained probabilistic frameworks: (i) \textit{PC-SNGP}, based on  Gaussian process~\cite{liu2020simple}, and (ii) \textit{PC-SNER}, based on Evidential Regression~\cite{amini2020deep}. Both frameworks enforce distance-preserving representations through spectral normalization and provide calibrated uncertainty estimates without sampling-based inference, thereby reducing computational cost. We further introduce a dynamically weighted physics-constrained loss to maintain a balance between data fidelity and physical consistency, enforced via introducing a unified physics-based degradation model that captures fatigue, wear, debris, and lubrication dynamics. We also introduce a distance-aware-coefficient (DAC) based on the Pearson correlation coefficient (PCC)~\cite{cohen2009pearson} to quantify sensitivity to distributional shifts.

Our contributions are as follows:
\begin{enumerate}
    \item Two sampling-free, distance-aware physics-constrained probabilistic frameworks (PC-SNGP and PC-SNER) that provide distance-aware uncertainty.
    \item A unified multicomponent physics-based degradation model incorporating fatigue, wear, debris, and lubrication dynamics.
    \item A dynamically weighted physics-constrained loss for adaptive training consistency and a distance-aware coefficient (DAC) metric, based on the Pearson correlation to quantify sensitivity to distributional shifts.
    \item Validation on the PRONOSTIA, XJTU-SY, and HUST bearing datasets, demonstrating strong out-of-distribution (OOD) generalization, robustness under cross-validation and adversarial scenarios.
\end{enumerate}

The remainder of the paper is organized as follows. Section~\ref{sec:literature} reviews the literature and related work. Section~\ref{sec:theory} presents the problem formulation, provides a theoretical background of all building blocks of developed frameworks. Section~\ref{sec:case_study} evaluate both frameworks. Section~\ref{sec:conclusion} concludes the paper with possible future work.

\section{Related Work}\label{sec:literature}

\textbf{Physics-based Prognostics Models:} Recent physics-based prognostic models have incorporated increasingly detailed representations of degradation mechanism. Guo et al.~\cite{guo2015fatigue} combined centrifugal expansion, thermal deformation, and Hertzian contact mechanics into a quasistatic stress distribution model. Gabrieli et al.~\cite{gabrielli2024physics} proposed the equivalent damaged volume concept to quantify defect severity via comparisons of measured and simulated vibration signals. Yang et al.~\cite{yang2025rolling} developed an electro-mechanically coupled digital twin that replicates contact forces and vibration under variable speed and fault conditions. Ohana et al.~\cite{ohana2025new} introduced a spall-progression framework that integrates dynamic response modeling with oil debris monitoring. While physically rigorous, these models do not inherently quantify predictive uncertainty, require precise parameterization by experts, degrade under modeling simplifications, and lack self-adaptation to evolving operating conditions, limiting their use in probabilistic prognostics. These limitation highlight the need for hybrid frameworks that seamlessly integrate physical consistency and rigor with probabilistic learning to enable uncertainty-aware prognostic. 

\textbf{Data-driven Probabilistic Prognostics Models:} Several data-driven probabilistic prognostics approaches have been developed. Jiang et al.~\cite{jiang2023remainingbnn} incorporated a CNN-LSTM architecture with Bayesian variational inference. Rivas et al.~\cite{rivas2022predictions} applied Bayes by Backprop to turbofan engine prognostics. Ochella et al.~\cite{ochella2024bayesian} utilized Monte Carlo dropout to jointly model aleatoric and epistemic uncertainty. Pan et al.~\cite{pan2024meta} proposed a meta-weighted network that combines diffusion-based uncertainty quantification with meta-learning for domain adaptation. Despite these advances, several limitations remain: (i) these methods require large volumes of high-quality labeled degradation data for effective training; (ii) their computational complexity increases significantly with model depth and the number of nonlinear parameters, complicating deployment on resource-constrained platforms; (iii) they often operate as black-box models that lack interpretability and explicit adherence to physical laws; and (iv) their uncertainty estimates are not distance aware and fail to reliably reflect distributional shifts between training and test data. These motivate physics-constrained learning to enforce physical interpretability while maintaining robust uncertainty quantification under distributional shifts,

\textbf{Physics-constrained Probabilistic Prognostics Models:} Physics-constrained probabilistic prognostics remains a nascent research area with representative approaches include: Bayesian PCMs (PC-BNNs)~\cite{linka2022bayesian}, Monte Carlo dropout PCMs~\cite{alhajeri2022physics}, and deep-ensemble PCMs~\cite{soibam2024inverse}. Despite incorporating physical constraints, these methods share several limitations: (i) high computational cost due to sampling-based inference procedure; (ii) reliance on large labeled datasets, which diminishes the data efficiency benefits of physics-constrained modeling; and (iii) lack of distance-aware uncertainty quantification. These limitations highlight sampling-free, distance-aware physics-constrained probabilistic frameworks that provide calibrated uncertainty estimates and robustness under distributional shifts. 

\section{Methods}\label{sec:theory}
In this section, we present the theoretical and methodological frameworks: (i) Problem formulation; (ii) Theory of Dynamic Physics-constrained Models (PCMs); (iii) Spectral Normalization; (iv) PC-SNGP; (v) PC-SNER; and (vi) Unified Physics-Based Degradation Model.

\subsection{Problem Formulation}

Let $\mathcal{X}=\{x_t\}_{t=1}^{T}$, with $x_t\in\mathbb{R}^{N_c}$, denote multivariate condition-monitoring time-series observations collected from $N_c$ channels. At each time step $t$, the model observes a temporal window:
\begin{equation}
\mathcal{X}=(x_1,x_2,\ldots,x_T),
\end{equation}
and PCM predicts a degradation indicator $y_t \in \mathbb{R}$. Given a training data sampled $\mathcal{D}=\{(x_i,y_i)\}_{i=1}^{N}$ from an in-domain region $\mathcal{X}_{\mathrm{IND}} \subset \mathcal{X}$, the goal is to learn a predictive distribution $p(y \mid x)$ that accurately estimate degradation while quantifying uncertainty. During testing, inputs may originate from unseen conditions or failure modes in an out-of-distribution region ($\mathcal{X}_{\mathrm{OOD}}=\mathcal{X}\setminus\mathcal{X}_{\mathrm{IND}}$), where the underlying conditional distribution may differ from training. Hence, a reliable model should be distance-aware, producing confident predictions near  $\mathcal{X}_{\mathrm{IND}}$ and increasing uncertainty as inputs deviate from the training manifold.

\subsection{Theory of Dynamic Physics-Constrained Models (PCMs)}\label{appendix:training}
PCMs extend standard neural network (\(\mathcal{NN}\)) models by embedding physics directly into the learning process~\cite{raissi2019physics}, encouraging convergence toward physically consistent solutions. This is typically achieved by augmenting the loss function with a physics-based term, resulting in a total loss of the form:
\begin{equation}
  \mathcal{L}_{\text{total}} = \mathcal{L}_{\text{data}} + \mathcal{L}_{\text{phys}}
\end{equation}
Here, \(\mathcal{L}_{\text{data}}\) quantifies the discrepancy between the model’s prediction and ground truth observations, whereas \(\mathcal{L}_{\text{phys}}\) penalizes deviation from established physical laws. This formulation helps mitigate the black-box nature of \(\mathcal{NN}\)s by enforcing physical consistency during training.

In the context of degradation estimation, certain feature indicators, such as those that exhibit monotonic trends over time, provide important cues. For example, an increase in specific feature values is expected to correspond to greater degradation. When such relationships are violated, the model’s prediction conflicts with physical expectations. To prevent this, \(\mathcal{L}_{\text{phys}}\) is computed using automatic differentiation, comparing the time derivative of the network's output \(D_{\text{phys}}\) with the output of the physical degradation model. However, degradation is inherently complex and nonlinear, and standard PCMs often struggle to maintain strict monotonicity in their prediction. To address this, we introduce dynamic weighting to the loss components, allowing the model to adaptively balance data fidelity and physical consistency during training. 
The weighting coefficients are determined from the variability of the input indicators and the physics-based degradation estimate. Specifically, the standard deviations of the indicator features, $\sigma_\mathcal{X}$, and the physics-based degradation estimate, $\sigma_{\mathrm{phys}}$, are used to construct the weight vector as

\begin{equation}
\begin{bmatrix}
w_{\mathrm{data}} \\
w_{\mathrm{phys}}
\end{bmatrix}
=
\mathrm{softmax}
\left(
\begin{bmatrix}
\sigma_\mathcal{X} \\
\sigma_{\mathrm{phys}}
\end{bmatrix}
\right),
\label{eq:dynamic_weights}
\end{equation}
where \(\sigma_{\text{phys}} = \text{StdDev}(D_{\text{phys}}) \) and \(\sigma_\mathcal{X} = \text{StdDev}(\mathcal{X}_{\text{IND}\sim\text{batch}})\). This formulation ensures that the loss contributions are adaptively scaled based on the variability of each component, promoting more stable and self-balanced optimization. The softmax normalization ensures that \(w_{\mathrm{data}} + w_{\mathrm{phys}} = 1, \quad w_{\mathrm{data}},\, w_{\mathrm{phys}} \ge 0.\). The combined loss becomes
\begin{equation}
\mathcal{L}_{\mathrm{total}}
=
w_{\mathrm{data}}\,\mathcal{L}_{\mathrm{data}}
+
w_{\mathrm{phys}}\,\mathcal{L}_{\mathrm{phys}}.
\label{eq:total_loss}
\end{equation}

\subsection{Spectral Normalization}
To make the PCM distance aware, it is essential to ensure that the hidden mapping \(h \) preserves distances; that is, the distance between inputs \(\|x - x'\|_X \) is approximately maintained in the hidden representation space \(\|h(x) - h(x')\|_H \). For networks with residual connections, this can be achieved by ensuring that each nonlinear residual block \(\{g_l\}_{l=1}^{L-1} \) is Lipschitz bound with a constant less than 1 \cite{liu2020simple}. Let the hidden mapping be defined as
\begin{equation}
    h = h_{L-1} \circ \dots \circ h_2 \circ h_1,
\end{equation}
with each residual block defined as \(h_l(x) = x + g_l(x) \). If every \(g_l \) is \(\alpha \)-Lipschitz for some \(0 < \alpha < 1 \), i.e.,
\begin{equation}
\|g_l(x) - g_l(x')\|_H \leq \alpha \|x - x'\|_X, \quad \forall x, x' \in \mathcal{X}_\text{IND},
\end{equation}
then the full mapping \(h \) satisfies the bi-Lipschitz condition:
\begin{equation}
(1 - \alpha)^{L-1} \|x - x'\|_X \leq \|h(x) - h(x')\|_H \leq (1 + \alpha)^{L-1} \|x - x'\|_X.
\label{eq:lips}
\end{equation}

To enforce the Lipschitz constraint, it suffices to control the spectral norm of the linear weights \(W_l \) in each block \(g_l(x) = W_l x + b_l \), since the Lipschitz constant of a linear transformation is upper bounded by its spectral norm: \(\|g_l\|_{\text{Lip}} \leq \|W_l\|_2 \). Therefore, we apply spectral normalization at each training step:
\begin{equation}
W_l \xleftarrow{}
\begin{cases}
\frac{c \cdot W_l}{\hat{\lambda}} & \text{if } \hat{\lambda} > c, \\
W_l & \text{otherwise},
\end{cases}
\label{eq:spectral_normalization}
\end{equation}
where \(\hat{\lambda} = \|W_l\|_2 \) is the estimated spectral norm and \(c < 1 \) is the norm-multiplier hyperparameter \cite{liu2020simple}.

\subsection{PC-SNGP: Physics-Constrained Spectral Normalization Gaussian Process}  \label{sec:pisngp}
To make the PCM output distance-aware, PC-SNGP replaces the dense output \(d: H \to \mathbb{R} \) using a Gaussian process (GP) with RBF kernel, where the posterior variance depends on the Euclidean distance between test and training hidden representations. Let \(\mathcal{D} = \{(x_i, y_i)\}_{i=1}^N \) be the training set, and \(h_i = h(x_i) \in \mathbb{R}^{D_{L-1}} \) be the penultimate layer outputs. The GP output vector \(\mathbf{g} = [g_1, \dots, g_N]^T \) follows \(\mathcal{MVN}(\mathbf{0}, K) \), where \(K \in \mathbb{R}^{N \times N} \) with entries $K_{ij} = \exp(-\gamma \| h_i - h_j \|_2^2)$, and \(\gamma \) is a tunable RBF kernel parameter. For scalability, the kernel is approximated using Random Fourier Features (RFF):
\begin{equation}
\Phi_i = \sqrt{\frac{2}{D_L}} \cos(W_L h_i + b_L)
\end{equation}
where \(W_L \in \mathbb{R}^{D_L \times D_{L-1}} \sim \mathcal{N}(0, I) \) and \(b_L \in \mathbb{R}^{D_L} \sim \mathcal{U}[0, 2\pi] \). The output becomes:
\begin{equation}
g(h_i) = \Phi_i^T \beta = \sqrt{\frac{2}{D_L}} \cos(W_L h_i + b_L)^T \beta,
\end{equation}
where \(\beta \in \mathbb{R}^{D_L} \) is a trainable weight vector. In a Bayesian linear regression setting, $\beta \sim \mathcal{MVN}(\hat{\beta}, \Sigma)$, where $\hat{\beta}$ is the predictive mean and posterior covariance: \(\Sigma = I_{D_L} - \Phi K^{-1} \Phi^T\) providing uncertainty estimates while ensuring distance-awareness and physical coherence \cite{liu2020simple}.

\textbf{Training Objective:} PC-SNGP can be trained end-to-end using a maximum likelihood estimation (MLE) objective. In the context of deterministic regression, it is assumed that each target value \(y_i \) is drawn independently and identically distributed (i.i.d.) from a Gaussian distribution parameterized by the model's predictive \(\hat{\beta}\) and \(\Sigma\), denoted collectively as \( \theta \). These parameters are learned by maximizing the likelihood of observing the target data, that is, by optimizing \( p(y_i \mid \theta) \). This is achieved by minimizing the negative log-likelihood (NLL), which for a Gaussian output distribution takes the form:
\begin{equation}
\mathcal{L}_{\text{MLE}} =  -\log p (y_i \mid \underbrace{\hat{\beta}, \Sigma}_{\theta}) = \frac{1}{2} \left[ \log \left(2\pi \, \mathrm{diag}(\Sigma) \right) + \frac{(y_i - \hat{\beta})^2}{\mathrm{diag}(\Sigma)} \right],
\label{eq:PC-SNGP_loss}
\end{equation}
where \( \mathrm{diag}(\Sigma) \) denotes the predictive variance obtained from the diagonal entries of the posterior covariance matrix. While this likelihood-based formulation effectively captures uncertainty.

\subsection{PC-SNER: Physics-Constrained Spectral Normalization Evidential Regression} \label{sec:piSNER}
To enable distance-aware evidential uncertainty quantification in PCMs, PC-SNER modifies the dense output layer to produce evidential priors. Using the same hidden representation \( h_i = h(x_i) \in \mathbb{R}^{D_{L-1}} \), the output layer now maps: \(d: h(x) \mapsto (\gamma, \nu, \alpha, \beta),\) which parameterizes a Normal-Inverse-Gamma (NIG) distribution over the mean (\(\mu\)) and variance (\(\sigma^2\)) of a Gaussian likelihood. The choice of the NIG distribution is motivated by the limitations of Prior Networks \cite{malinin2019reverse, malinin2018predictive}, which place Dirichlet Priors \cite{sethuraman1994constructive} over discrete classification predictions and require OOD data or fixed prior regularization, whereas NIG enables joint modeling of aleatoric and epistemic uncertainty in continuous regression tasks \cite{amini2020deep}. The NIG distribution becomes:
\begin{equation}
p(\mu, \sigma^2 \mid
\underbrace{\gamma, \nu, \alpha, \beta}_{\theta}) =
\frac{\beta^\alpha \sqrt{\nu}}{\Gamma(\alpha)\sqrt{2\pi \sigma^2}}
\left( \frac{1}{\sigma^2} \right)^{\alpha + 1}
\exp\left( -\frac{2\beta + \nu(\gamma - \mu)^2}{2\sigma^2} \right)
\end{equation}

This induces the marginal distribution:
\begin{equation}
\mu \sim \mathcal{N}(\gamma, \frac{\sigma^2}{\nu}), \quad \sigma^2 \sim \Gamma^{-1}(\alpha, \beta),
\end{equation}
yielding a Student-\( t \) distribution over the predictive output \( y \). The predictive mean is \( \mathbb{E}[y] = \gamma \), and the predictive variance is
\begin{equation}
\text{Var}[y] =
\underbrace{\frac{\beta}{\alpha - 1}}_{\text{aleatoric}} +
\underbrace{\frac{\beta}{\nu(\alpha - 1)}}_{\text{epistemic}} =
\frac{\beta(1 + \nu)}{\nu(\alpha - 1)}
\end{equation}
capturing both aleatoric and epistemic uncertainty. The output corresponding to the mean $\gamma$ uses a linear activation, while the remaining parameters $\nu, \alpha, \beta$ are passed through a \textit{softplus} activation to ensure positivity. Additionally, a small positive shift is applied to $\alpha$ ($\alpha \gets \text{softplus}(z) + 1$) to ensure the condition $\alpha > 1$, which is required for the existence of the first moment of the NIG distribution.

\textbf{Training Objective:} PC-SNER is trained end-to-end using a two-term objective designed to both maximize data fit and penalize unwarranted certainty. The first term of the loss is the negative log marginal likelihood under the Student-$t$ distribution. Given a training set $\mathcal{D} = \{(x_i, y_i)\}_{i=1}^N$, and denoting the network’s predicted parameters for the $i$-th sample as $(\gamma_i, \nu_i, \alpha_i, \beta_i)$, the negative log-evidence is defined as
\begin{equation}
\mathcal{L}_{\text{NLE}} = \frac{1}{2} \log\left( \frac{\pi}{\nu_i} \right) - \alpha_i \log(\Omega_i) + \left( \alpha_i + \frac{1}{2} \right) \log\left[(y_i - \gamma_i)^2 \nu_i + \Omega_i \right] + \log\left( \frac{\Gamma(\alpha_i)}{\Gamma(\alpha_i + \frac{1}{2})} \right)
\end{equation}
where $\Omega_i = 2\beta_i(1 + \nu_i)$. This term encourages the network to produce evidence that is consistent with the data by maximizing the marginal likelihood of each target value under the induced predictive distribution. To prevent the network from generating overconfident predictions on erroneous inputs, a regularization term is introduced that penalizes the amount of evidence in proportion to the prediction error. Drawing from the interpretation of the NIG parameters as virtual observations, the total evidence for the $i$-th sample is given by $\Phi_i = 2\nu_i + \alpha_i$. The regularization loss ($\mathcal{L}_{\text{Reg}}$) is therefore defined as
\begin{equation}
\mathcal{L}_{\text{Reg}} = |y_i - \gamma_i| \cdot \Phi_i = |y_i - \gamma_i| \cdot (2\nu_i + \alpha_i)
\end{equation}
This regularizer discourages overconfident estimates when prediction errors are large, implicitly encouraging the model to inflate uncertainty in regions of the input space where the prediction is uncertain or out-of-distribution. The total loss ($\mathcal{L}_{\text{ER}}$) for each training sample is a weighted sum of the two components:
\begin{equation}
\mathcal{L}_{\text{ER}} = \mathcal{L}_{\text{NLE}} + \lambda \cdot \mathcal{L}_{\text{Reg}}
\label{eq:PC-SNER_loss}
\end{equation}
where $\lambda \in \mathbb{R}_+$ is a hyperparameter controlling the trade-off between fitting the data and enforcing conservative uncertainty estimates \cite{amini2020deep}.

\begin{wrapfigure}{r}{0.42\columnwidth}
\centering
\vspace{-3mm}
\centering
\includegraphics[width=0.4\textwidth]{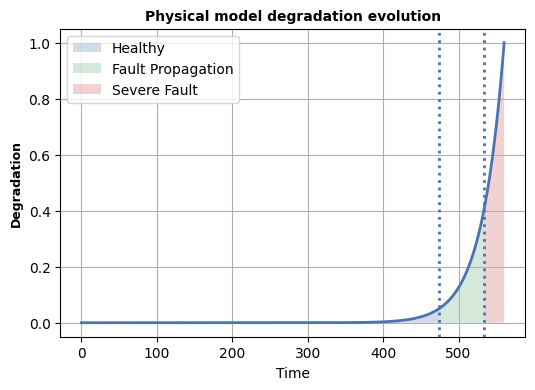}
\caption{Fault evolution of physical model.}
\label{fig:fault}
\end{wrapfigure}

\subsection{Unified Physics-Based Degradation Model}
We develop a unified degradation model for REBs that captures fatigue, abrasive wear, lubrication breakdown, and stochastic contamination within a single coupled framework. Degradation is represented through coupled ordinary and stochastic differential equations, providing a coherent description of nonlinear progression over time. The total degradation is expressed as
\begin{equation}
\dot{D}_{\text{coupled}} = \dot{D}_F + \gamma_w \dot{D}_W + \zeta_L \dot{D}_O.
\label{eq:degradation_physics}
\end{equation}
where \(D_F\) is fatigue degradation, \(D_W\) is the wear volume including surface roughness effects, and \(D_O\) represents lubrication degradation with thermal feedback. This formulation successfully captures the nonlinear dynamics of degradation and clearly describes three stages of bearing life: (i) Healthy, (ii) Fault Progression, and (iii) Severe Fault. The evolution of these faults is illustrated in Figure~\ref{fig:fault}. A full derivation and detailed explanation are provided in Appendix~\ref{appendix:physical}.

\section{Evaluation}\label{sec:case_study}

\subsection{Data, Curation, Evaluation Metrics, Training Strategy}
We employed three publicly available datasets to evaluate the developed frameworks: (i) PRONOSTIA \cite{nectoux2012pronostia}, (ii) XJTU-SY \cite{wang2018xjtu}, and (iii) HUST \cite{thuan2023hust}. Training is performed on the last four bearings (4 to 7) from \textit{Condition 1} from the PRONOSTIA dataset, while cross-validation experiments in zero-shot manner are conducted on the other two datasets. Detailed descriptions and training/testing splits of the datasets are provided in Appendix~\ref{appendix:description}. All of the datasets consist of 1-D nonstationary vibrational signals that must be preprocessed to extract meaningful degradation patterns. We adopted the preprocessing methodology proposed in author's previous work \cite{razzaq2025carle}. Detailed information is provided in Appendix~\ref{appendix:curation}. The performance is evaluated using the mean-squared-error (MSE), mean-absolute error-(MAE), Score~\cite{nectoux2012pronostia}, and a novel distance-aware-coefficient (DAC). Detailed definitions of these evaluation metrics are provided in Appendix~\ref{appendix:metrics}.

\begin{wrapfigure}{r}{0.53\textwidth}
\vspace{-8mm}
\begin{minipage}{0.51\textwidth}
\begin{algorithm}[H]
\small
\caption{Training Algorithm of PCM}
\label{algo:train_step}
\begin{algorithmic}
\Require Input batch: $\mathcal{X}_{\text{batch}}, y_{\text{true}}$, Physics inputs: $\text{Load}, \text{RPM}, t_{\text{batch}}, T_{\text{batch}}$,  Model $\mathcal{M_\theta}$ with parameters $\theta$, Adam optimizer $\mathcal{O}$, Physical model $k_{\text{phys}}$, \(\mathcal{L}_{\text{MLE}}\), \(\mathcal{L}_{\text{ER}}\)
\State Data prediction: $\hat{y} \gets \mathcal{M_\theta}(X_{\text{batch}})$
\State Calculate data loss: $\mathcal{L}_{\text{data}} \gets \mathcal{L}_{\text{MLE}} \text{ || } \mathcal{L}_{\text{ER}} (y, \hat{y})$
\State Physical input: $\mathcal{X}_{\text{phys}} \gets \text{concat}(\mathbf{0}_{N \times 14}, t_{\text{batch}}, T_{\text{batch}})$
\State Enable gradient tracking for $\mathcal{X}_{\text{phys}}$
\State Physical prediction: $D_{\text{pred}} \gets \mathcal{M_\theta}(\mathcal{X}_{\text{phys}})$
\State Automatic differentiation: $\frac{\partial D}{\partial t} \gets \nabla_{\mathcal{X}_{\text{phys}}} D_{\text{pred}}[:, 14]$
\State Physical output: $D'_{\text{phys}} \gets D_{\text{coupled}}(t_{\text{batch}}, \text{Load}, \text{RPM}, T_{\text{batch}})$
\State Physical loss: $ \mathcal{L}_{\text{phys}} \gets \mathcal{L}_{\text{MLE}} \text{ || } \mathcal{L}_{\text{ER}} \left(\frac{\partial D}{\partial t}, D'_{\text{phys}}\right)$
\State Total loss: $\mathcal{L}_{\text{total}} \gets w_{\text{data}} \mathcal{L}_{\text{data}} +w_{\text{phys}}  \mathcal{L}_{\text{phys}}$
\State Compute gradients: $\nabla_\theta \mathcal{L}_{\text{total}}$
\State U$\rho_d$ate parameters: $\theta \gets \mathcal{O}(\theta, \nabla_\theta \mathcal{L}_{\text{total}})$
\State \Return $\mathcal{L}_{\text{total}}, \mathcal{L}_{\text{data}}, \mathcal{L}_{\text{phys}}$
\end{algorithmic}
\end{algorithm}
\end{minipage}
\vspace{-2mm}
\end{wrapfigure}

\subsubsection{Training Algorithm of the PCMs}
Algorithm~\ref{algo:train_step} outlines the custom training step of the PCM. At each training step, the model processes two complementary input streams: (1) the TFR representation, \(\mathcal{X}_{\text{batch} \sim \text{TFR}}\), and (2) structured inputs that encode physical parameters, \(\mathcal{X}_{\text{batch} \sim \text{phys}}\) (see Table \ref{tab:physical} for physical variables). The model first computes the prediction for the data-driven stream (\(\hat{y}\)) and then compares the predictions to the ground -truth labels (\(y\)) to compute loss using Eqn.~\ref{eq:PC-SNGP_loss} (for PC-SNGP) or Eqn.~\ref{eq:PC-SNER_loss} (for PC-SNER). Moreover, the physics-constrained pathway introduces domain knowledge by evaluating how well the model’s behavior aligns with known physical laws. This is done by using automatic differentiation to compute the gradient (\(\frac{\partial D}{\partial t}\)) of the physical prediction \(D_{\text{pred}}\) with respect to (\(t\)), which is then compared to the expected output (\(D'_{\text{phys}}\)) from the theoretical model (\(D_{\text{coupled}}\)) from Eqn. \ref{eq:degradation_physics}. A composite loss \(\mathcal{L}_{\text{total}}\) function combines both the \(\mathcal{L}_{\text{data}}\) and \(\mathcal{L}_{\text{phys}}\), weighted by parameters derived in Eqn.~\ref{eq:total_loss}, balancing empirical accuracy and physical consistency. During backpropagation, gradients from both loss components guide parameter updates, enabling a multiobjective optimization process. This dual-stream architecture helps ensure that the learned representations capture patterns in the data and follow the core principles of the underlying physical system.

\subsection{Results}
The baseline \(\mathcal{NN}\) comprises six MLP hidden layers with a neuron configuration [32, 32, 64, 64, 32, 32] and ReLU activations. We compare both frameworks against MC-dropout~(PC-MC) \cite{alhajeri2022physics} and Deep-Ensemble~(PC-DE) \cite{lakshminarayanan2017simple}. For PC-SNGP, we consider the impact of the RBF kernel parameter \(\gamma\) in the covariance function, three values are tested: 0.5, 1.0, and 2.0. For PC-SNER, we investigate the influence of the regularization parameter \(\lambda\), considering three values: 0.2, 0.5, and 1.0. For PC-MC, we explored the effect of applying dropout during inference, using three rates: 0.1, 0.3, and 0.5 after each hidden layer. For PC-DE, we consider the impact of the number of deep models: $N_m\in [10, 20, 30]$. For all tests, we visualized \((\mu\)) along with respective uncertainty (aleatoric, epistemic, or both) bands extending up to \(2\sigma\), as well as contour plots, to facilitate comparison and analysis of the predictive behavior and uncertainty quantification.

\begin{figure}[ht!]
\centering
\begin{minipage}[c]{0.49\textwidth}
\centering
\includegraphics[width=0.98\linewidth]{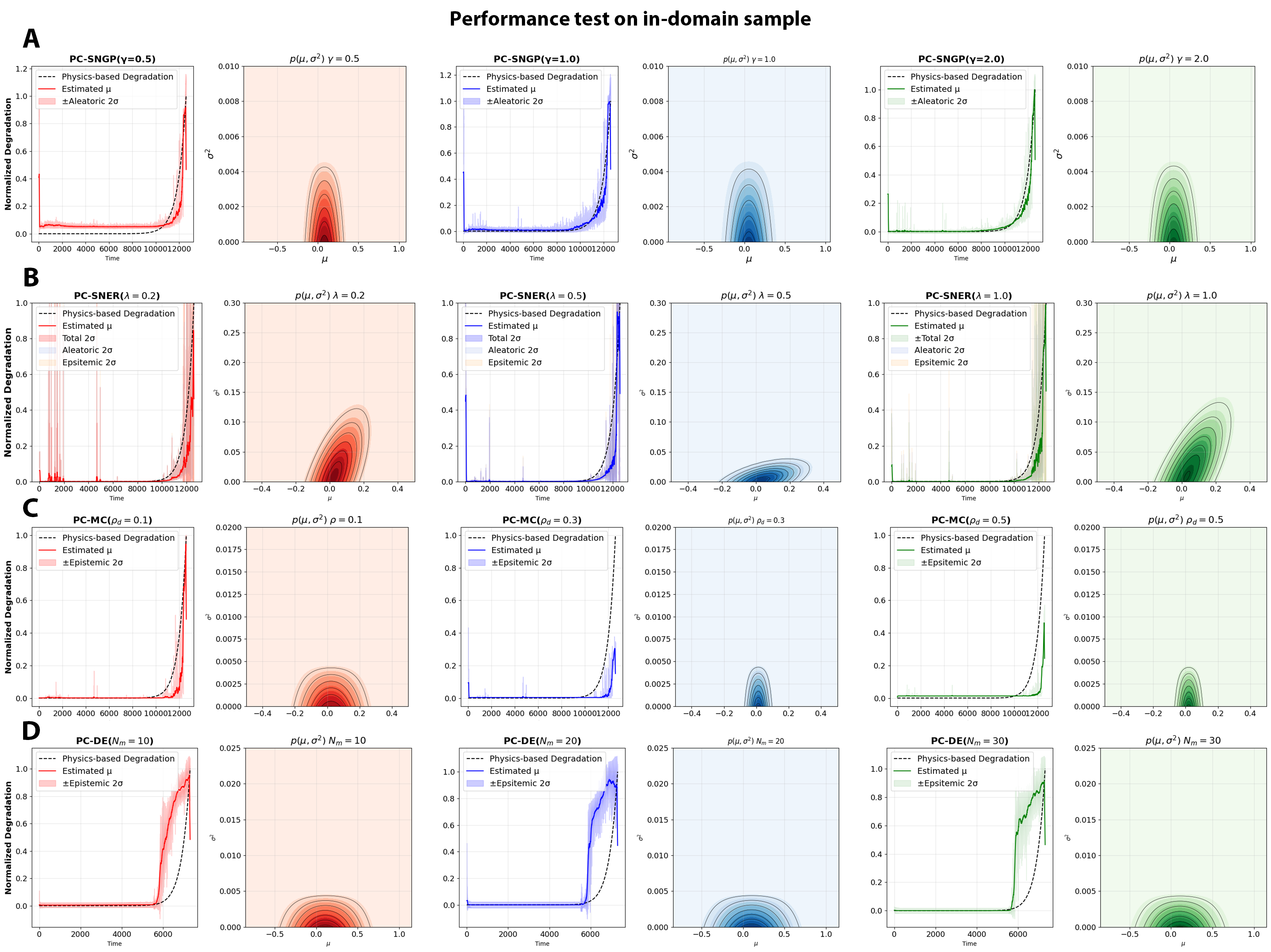}
\caption{Results of the variation in the respective parameters for \textit{Bearing~5} \(\in \mathcal{X}_{\text{IND}}\) sample.}
\label{fig:ind}
\end{minipage}
\hfill
\begin{minipage}[c]{0.49\textwidth}
\captionsetup{type=table}
\caption{Summary of metric of performance test under \textit{Bearing~5 }\(\in \mathcal{X}_{\text{IND}}\) sample}
\label{tab:metric_ind}
\small
\centering
\resizebox{0.95\linewidth}{!}{%
\begin{tabular}{@{}lSSSS@{}}
\toprule
\rowcolor{gray!20}
\textbf{Model} & \textbf{MSE} & \textbf{MAE} & \textbf{Score} & \textbf{DAC} \\
\midrule
PC-SNGP($\gamma=0.5$) & 0.0066 & 0.0625 & 74.07 & 0.1234 \\
PC-SNGP($\gamma=1.0$) & 0.0031 & 0.0237 & 26.86 & 0.3188 \\
PC-SNGP($\gamma=2.0$) & 0.0021 & 0.0164 & 17.35 & 0.2886 \\
PC-SNER($\lambda=0.2$) & 0.0030 & 0.0493 & 49.21 & 0.0025 \\
PC-SNER($\lambda=0.5$) & 0.0007 & 0.0182 & 19.32 & 0.4788 \\
PC-SNER($\lambda=1.0$) & 0.0014 & 0.0208 & 23.12 & 0.3216 \\
PC-MC($p_d=0.1$) & 0.0045 & 0.0306 & 36.51 & \text{-} \\
PC-MC($p_d=0.3$) & 0.0303 & 0.1143 & 145.91 & \text{-} \\
PC-MC($p_d=0.5$) & 0.549 & 0.5676 & 436.13 & \text{-} \\
PC-DE($N_m=10$) & 0.0012 & 0.0162 & 16.95 & \text{-} \\
PC-DE($N_m=20$) & 0.0009 & 0.0226 & 22.19 & \text{-} \\
PC-DE($N_m=30$) & 0.0012 & 0.0232 & 23.03 & \text{-} \\
\bottomrule
\end{tabular}}
\begin{minipage}{0.95\linewidth}
\footnotesize
\textbf{Note:} (-) indicate no distance-awareness.
\end{minipage}
\end{minipage}
\end{figure}

\subsubsection{Performance on \(\mathcal{X}_{\text{IND}}\) sample}
To analyze the performance on \(\mathcal{X}_{\text{IND}}\), we selected \textit{Bearing~5} from the training manifold. Figure~\ref{fig:ind} presents the qualitative results and metrics are reported in Table~\ref{tab:metric_ind}. For PC-SNGP (Figure~\ref{fig:ind}A), increasing the RBF kernel parameter $\gamma$ progressively tightens uncertainty bounds and improves the predictive accuracy; $\gamma=1.0$ achieves the strongest distance-aware balancer (MSE=0.0031, DAC=0.3188), whereas $\gamma=2.0$ yields the lowest error (MSE=0.0021) at the cost of DAC degradation. For PC-SNER (Figure~\ref{fig:ind}B), $\lambda=0.5$ achieves the best overall performance (MSE=0.0007; MAE=0.0182; Score=19.33; DAC=0.4788), with a higher $\lambda$ introducing progressive calibration degradation. MC-dropout (Figure~\ref{fig:ind}C) decreases sharply above $\rho_d=0.1$, with the MSE increasing to 0.549 and $\rho_d=0.5$, making the model unreliable. PC-DE (Figure~\ref{fig:ind}D) achieves competitive predictive accuracy at $N_m=20$ (MSE=0.0009) but provides no distance-aware calibration.

\begin{figure}[ht!]
\centering
\begin{minipage}[c]{0.49\textwidth}
\centering
\includegraphics[width=0.98\linewidth]{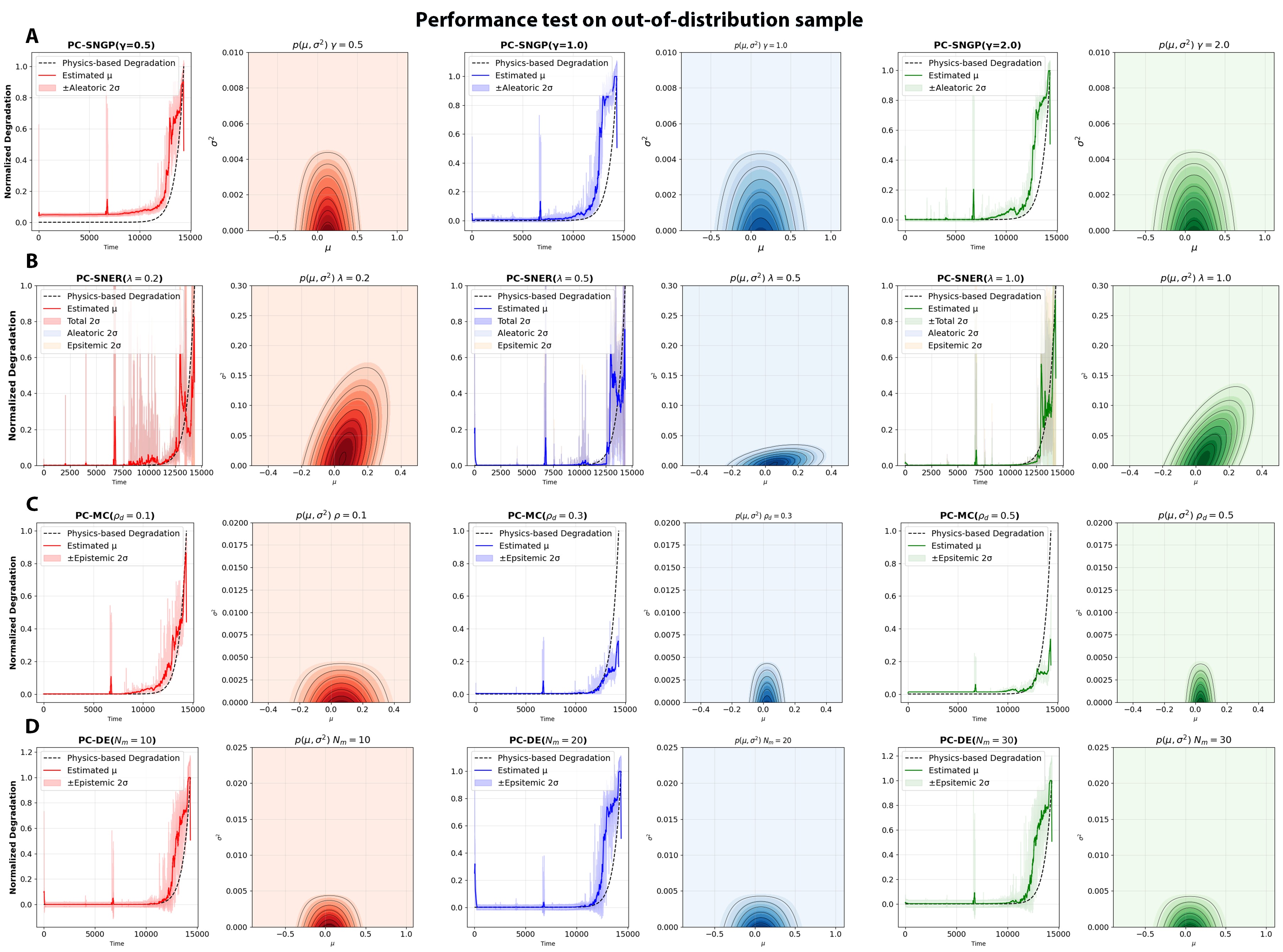}
\caption{Results of the variation in the respective parameters for \textit{Bearing~1} \(\in \mathcal{X}_{\text{OOD}}\) sample.}
\label{fig:ood}
\end{minipage}
\hfill
\begin{minipage}[c]{0.49\textwidth}
\captionsetup{type=table}
\caption{Summary of metric generalization test under \textit{Bearing~1 }\( \in \mathcal{X}_{\text{OOD}}\) sample.}
\label{tab:metric_ood}
\centering
\resizebox{0.95\textwidth}{!}{%
\begin{tabular}{@{}lSSSS@{}}
\toprule
\rowcolor{gray!20}
\textbf{Model} & \textbf{MSE} & \textbf{MAE} & \textbf{Score} & \textbf{DAC} \\
\midrule
PC-SNGP($\gamma=0.5$) & 0.0129 & 0.0811 & 117.08 & 0.3304 \\
PC-SNGP($\gamma=1.0$) & 0.0116 & 0.0766 & 112.08 & 0.4098 \\
PC-SNGP($\gamma=2.0$) & 0.0206 & 0.0632 & 92.04 & 0.3929 \\
PC-SNER($\lambda=0.2$) & 0.0018 & 0.0321 & 36.69 & 0.4123 \\
PC-SNER($\lambda=0.5$) & 0.0016 & 0.0281 & 39.54 & 0.4744 \\
PC-SNER($\lambda=1.0$) & 0.0019 & 0.0264 & 37.19 & 0.4244 \\
PC-MC($p_d=0.1$) & 0.0068 & 0.0472 & 61.15 & \text{-} \\
PC-MC($p_d=0.3$) & 0.0337 & 0.1068 & 155.52 & \text{-} \\
PC-MC($p_d=0.5$) & 0.0714 & 0.1998 & 291.65 & \text{-} \\
PC-DE($N_m=10$) & 0.0017 & 0.0348 & 38.49 & \text{-} \\
PC-DE($N_m=20$) & 0.0021 & 0.0369 & 40.98 & \text{-} \\
PC-DE($N_m=30$) & 0.0012 & 0.0261 & 29.69 & \text{-} \\
\bottomrule
\end{tabular}}
\begin{minipage}{0.95\linewidth}
\footnotesize
\textbf{Note:} (-) indicate no distance-awareness.
\end{minipage}
\end{minipage}
\end{figure}

\subsubsection{Performance on \(\mathcal{X}_{\text{OOD}}\) sample}
To analyze the performance on \(\mathcal{X}_{\text{OOD}}\), we selected \textit{Bearing~1} from \textit{Conditon~1} of the PRONOSTIA dataset. Figure~\ref{fig:ood} presents qualitative predictive outputs with uncertainty contours and quantitative metrics are reported in Table~\ref{tab:metric_ood}. For PC-SNGP (Figure~\ref{fig:ood}A), $\gamma=0.5$ achieves the prediction error (MSE=0.0129; Score=117.09), whereas $\gamma=1.0$ yields best MSE (0.0116) and DAC of 0.4098. PC-SNER (Figure~\ref{fig:ood}B) demonstrated superior OOD robustness at $\lambda=0.5$, with MSE=0.0016, Score=39.54, and DAC=0.4744. PC-MC (Figure~\ref{fig:ood}C) decreases substantially as instability emerges at $\rho_d=0.5$ (MSE=0.0714). PC-DE (Figure~\ref{fig:ood}D) again achieves competitive accuracy at $N_m=30$ (MSE=0.0012; Score=29.69) but offers no distance-aware uncertainty. 

\vfill
\pagebreak

\begin{wrapfigure}{r}{0.55\columnwidth}
\centering
\includegraphics[width=0.54\columnwidth]{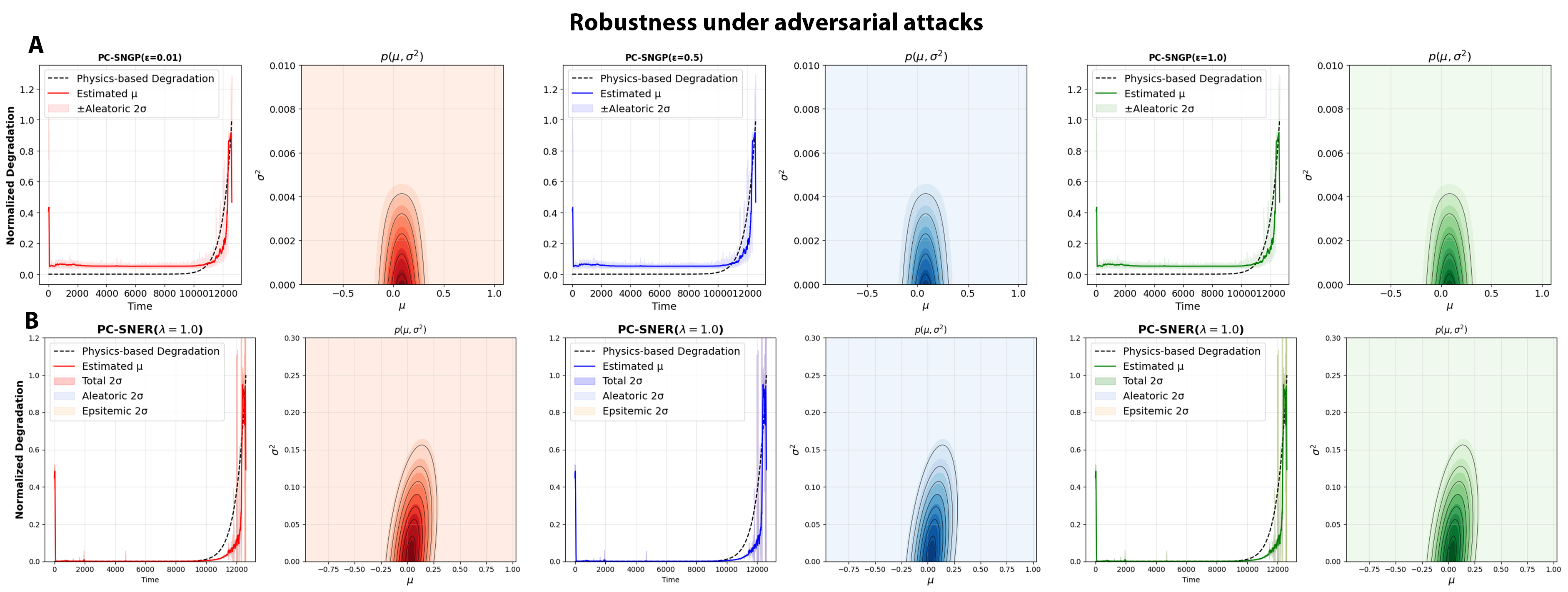}
\caption{Results of the variation in the perturbation magnitude \(\epsilon\) for \textit{Bearing~1} \(\in \mathcal{X}_{\text{OOD}}\) sample.}
\label{fig:adv}
\vspace{-3mm}
\end{wrapfigure}

\textbf{Robustness under adversarial attacks}: Adversarial resilience is evaluated by exposing $\mathcal{X}_{\mathrm{OOD}}$ samples to the Fast Gradient Sign Method (FGSM)~\cite{liu2019sensitivity} perturbation at magnitude $\epsilon \in {0.01, 0.5, 1.0}$ using the best-performing configuration: PC-SNGP ($\gamma=1.0$) and PC-SNER ($\lambda=0.5$). Figure~\ref{fig:adv} presents predictive trajectories with uncertainty contours. PC-SNGP~(Figure~\ref{fig:adv}A) maintains stable predictive means and tightly concentrated contour distributions across all $\epsilon$ values with no substantial variance inflation. Figure~\ref{fig:adv}B confirm analogous resilience in the PC-SNER, where both the aleatoric and epistemic uncertainty bands remain well-bound despite increasing perturbation magnitude. The observed robustness is attributed to two primary features: (i) Spectral Normalization improves stability by constraining the model’s Lipschitz constant; and (ii) the physics-constrained loss function, which imposes physical constraints that act as regularization, thereby suppressing overfitting and enhancing generalization and effectively.

\begin{figure}[ht!]
\centering
\begin{minipage}[c]{0.49\textwidth}
\centering
\includegraphics[width=0.9\textwidth]{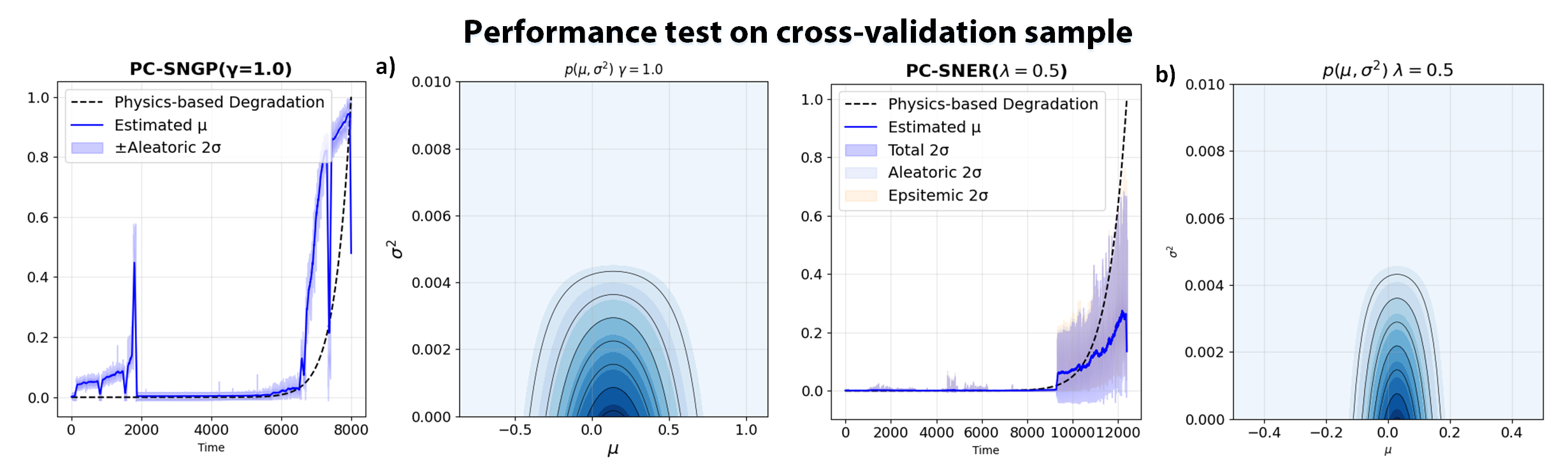}
\caption{Results of the \(\mathcal{X}_{\text{CV}}\) samples.}
\label{fig:cv}

\end{minipage}
\hfill
\begin{minipage}[c]{0.49\textwidth}
\captionsetup{type=table}
\caption{Summary of metrics cross-validation test under \(\mathcal{X}_{\text{CV}}\) sample.}
\label{tab:cross_val}
\vspace{1mm}
\centering
\resizebox{0.85\textwidth}{!}{%
\begin{tabular}{@{}l l S S S S@{}}
\toprule
\rowcolor{gray!20}
\textbf{Dataset} & \textbf{Model} & \textbf{MSE} & \textbf{MAE} & \textbf{Score} & \textbf{DAC} \\
\midrule
\textbf{XJTU-SY} & PC-SNGP & 0.0319 & 0.0851 & 69.02 & 0.34 \\
\textbf{HUST} & PC-SNER & 0.0019 & 0.0337 & 37.54 & 0.12 \\
\bottomrule
\end{tabular}}
\end{minipage}
\end{figure}

\subsubsection{Performance on \(\mathcal{X}_{\text{CV}}\) samples}
Cross-validation is essential in degradation estimation because it verifies that the learning can be generalized reliably to unseen data and novel degradation patterns. We conducted cross-validation experiments for PC-SNGP using \textit{Bearing~1} from \textit{Condition~1} of the XJTU-SY dataset, whereas cross-validation for PC-SNER was performed using \textit{Bearing~1} from \textit{Condition~1} of the HUST dataset. The qualitative results are shown in Figure~\ref{fig:cv}, and the quantitative metrics are summarized in Table~\ref{tab:cross_val}. Both models demonstrate strong performance while preserving distance awareness. PC-SNGP achieves an MSE of 0.0319, MAE of 0.0851, Score of 69.02, and DAC of 0.34. PC-SNER achieves MSE of 0.0019, MAE of 0.0337, Score of 37.54, and DAC of 0.12. These results indicate the strong cross-validation ability and robustness of the developed approach.

\begin{figure}[ht!]
\centering
\begin{minipage}[c]{0.49\textwidth}
\centering
\includegraphics[width=0.9\textwidth]{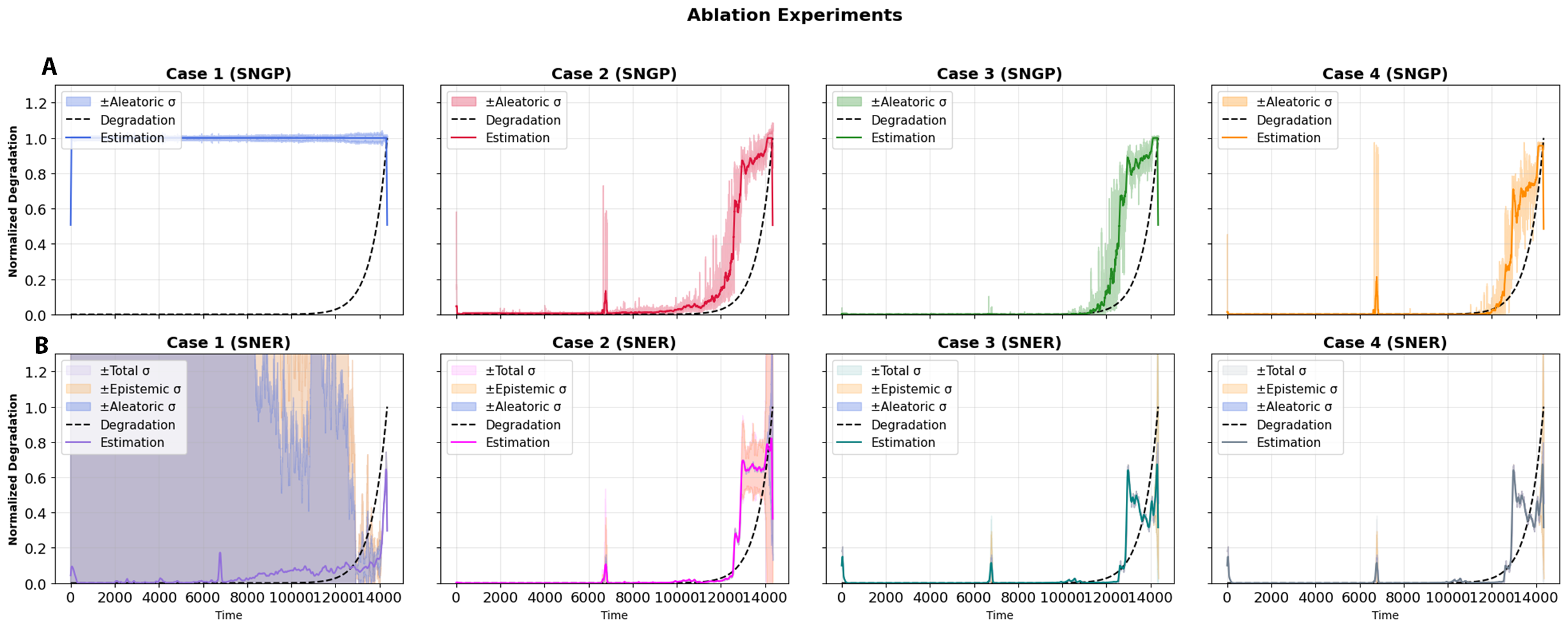}
\caption{Ablation Experiment results for \(\mathcal{X}_{\text{OOD}}\) sample.}
\label{fig:ablation}

\end{minipage}
\hfill
\begin{minipage}[c]{0.49\textwidth}
\captionsetup{type=table}
\caption{Summary of metrics ablation tests under \(\mathcal{X}_{\text{OOD}}\) sample.}
\label{tab:ablation_results}
\centering
\resizebox{0.9\textwidth}{!}{%
\begin{tabular}{@{}l l S S S S@{}}
\toprule
\rowcolor{gray!20}
\textbf{Model} & \textbf{Case} & \textbf{MSE} & \textbf{MAE} & \textbf{Score} & \textbf{DAC} \\
\midrule
\multirow{4}{*}{\textbf{PC-SNGP}} 
& Case 1  & 0.9223 & 0.9480 & 1428.79 & 0.3379 \\
& Case 2  & 0.0329 & 0.0762 & 111.77 & 0.5294 \\
& Case 3  & 0.0332 & 0.0668 & 98.01 & \text{-} \\
& Case 4  & 0.0116 & 0.0760 & 112.08 & 0.4098 \\
\midrule
\multirow{4}{*}{\textbf{PC-SNER}} 
& Case 1  & 0.0565 & 0.2021 & 252.63 & 0.0481 \\
& Case 2  & 0.0021 & 0.0295 & 31.48 & 0.2765 \\
& Case 3  & 0.0032 & 0.0366 & 42.59 & \text{-} \\
& Case 4  & 0.0016 & 0.0281 & 39.54 & 0.4744 \\
\bottomrule
\end{tabular}}
\begin{minipage}{0.95\linewidth}
\footnotesize
\textbf{Note:} (-) indicate no distance-awareness.
\end{minipage}
\end{minipage}
\end{figure}

\subsubsection{Ablation Study}
We conduct ablation experiments to analyze and verify the contribution of each building block of both frameworks. Four configurations are analyzed: Case 1 (physics-guidance removed), Case 2 (dynamic-weights removed), Case 3 (spectral normalization removed), and Case 4 (complete model). The qualitative results for both the PC-SNGP and the PC-SNER across all the cases are presented in Figure~\ref{fig:ablation}, with the quantitative metrics reported in Table~\ref{tab:ablation_results}. For PC-SNGP (Figure~\ref{fig:ablation}A), Case 1 fails to generalize because of underfitting in the absence of physical constraints (MSE=0.9223; Score=1428.79). Case 2 partially improve accuracy (MSE=0.0329). Case 3 yields a functionally identical accuracy (MSE=0.0332) but makes the DAC undefined, confirming that spectral normalization is an exclusive enabler of distance-aware latent geometry. Case 4 achieves optimal performance (MSE=0.0116, DAC =0.4098). For PC-SNER (Figure~\ref{fig:ablation}B), a similar pattern observed with Case 4 yielding the best metrics (MSE=0.0016, MAE=0.0281, DAC=0.4744), and all incomplete configurations lose distance awareness. These results establish that physical guidance, dynamic weighting, and spectral normalization are complementary and collectively essential for reliable uncertainty quantification.

\section{Conclusion}\label{sec:conclusion}
In this work, we developed two sampling-free, distance-aware physics-constrained probabilistic frameworks: (i) PC-SNGP; and (ii) PC-SNER for industrial prognostics. Both frameworks apply spectral normalization to hidden layer weights to enforce bi-Lipschitz distance-preserving representation from input to latent space. \textit{PC-SNGP} replaces the output layer with a Gaussian Process, where posterior variance increases as the test input moves away from the training manifold. \textit{PC-SNER} modifies the output layer to produce Normal-Inverse-Gamma (NIG) evidential parameters for distance-preserving estimation. A dynamically weighted physics-constrained loss is also developed that adaptively balances the data fidelity with physical consistency, enforced via a unified physics-based degradation model that incorporates fatigue, wear, lubrication, and debris accumulation to capture nonlinear degradation dynamics. To quantitatively assess sensitivity to distributional shifts, we developed a distance-aware coefficient (DAC) metric. We validate both frameworks on degradation estimation of rolling-element bearings (REBs) using PRONOSTIA, HUST, and XJTU-SY datasets. Experimental results demonstrated that both frameworks not only improve point-prediction accuracy but also preserve sensitivity to distributional shifts in uncertainty quantification against MC-dropout and Deep-Ensemble baselines. Both frameworks also maintained auditable performance in cross-validation as well as robustness under extreme adversarial perturbations.

\subsection{Future Work}
Future work may explore hybrid architectures that combine the complementary strengths of Gaussian Process and evidential approaches, potentially improving both uncertainty calibration and predictive performance. Furthermore, extending validation beyond rolling-element bearings and investigating more advanced latent-space distance measures and physics-based degradation models could provide additional insights into improving robustness and generalization under complex real-world distributional shifts.

\section*{Acknowledgments}
This research was supported by the CAS-ANSO Scholarship. We acknowledge the intellectual and material contributions of the University of Science and Technology of China (USTC) and the Alliance of International Science Organizations (ANSO). AI/LLM tools were used to polish the writing of the manuscript under strict human supervision.

\section*{Ethics Approval}
This study was conducted in accordance with ethical standards.

\section*{Data availability}
The code for reproducibility is available at \url{https://github.com/itxwaleedrazzaq/uqpcnn_rul}.

\section*{Authors Contribution}
\textbf{Waleed Razzaq:} Conceptualization, Methodology, Data Curation, Writing- Original draft preparation. \textbf{Yun-Bo Zhao}: Supervision, Writing- Reviewing.

\section*{Conflict of interest}
The authors declare that they have no known competing financial interests or personal relationships that could have appeared to influence the work reported in this paper.

\section*{Funding}
The research did not receive any funding from any organization.

\section*{Human/Animal Participation}
No human or animal participation is involved in this research.

\bibliographystyle{unsrt}
\bibliography{references}

\appendix
\section*{Appendix}
\section{Physics-based Degradation Model}\label{appendix:physical}
We develop a unified degradation model for REBs that captures fatigue, abrasive wear, lubrication breakdown, and stochastic contamination within a single coupled framework. Degradation is represented through coupled ordinary and stochastic differential equations, providing a coherent description of nonlinear progression over time. The model variables are:  
\begin{itemize}
  \item $P$: radial load on the bearing (N).  
  \item $n$: rotational speed (RPM), $\omega$: angular speed (rad/s).  
  \item $C_{\mathrm{load}}$: nominal dynamic load rating (N)
  \item $C_{\mathrm{eff}}(t)$: time-varying effective load rating (N) coupling all degradation mechanisms.  
  \item $D_W(t)$: accumulated wear volume (m$^3$)
  \item $V_d(t)$: equivalent damaged volume for fatigue (m$^3$).  
  \item $R(t)$: surface roughness metric.  
  \item $O(t)$: lubricant oxidation state (dimensionless index, e.g., normalized TAN)
  \item $\nu(t)$: lubricant kinematic viscosity ($m^2/s$).  
  \item $C_{\mathrm{debris}}(t)$: debris concentration (particles per unit volume).  
  \item $\mathcal{W}_t$: standard Wiener process representing stochastic contamination.  
  \item $D_{\text{coupled}}(t)$: cumulative damage index (dimensionless).  
\end{itemize}

\subsection{Components model and Physical Rationale}

\subsubsection{Fatigue: Hertzian contact, Lundberg--Palmgren scaling, and EDV}
Fatigue in REBs arises from subsurface shear stresses induced by repeated Hertzian contact, where material failure reflects the gradual accumulation of microstructural damage. Classical fatigue models capture this process through probabilistic cycle counting~\cite{zahavi2019fatigue}, but neglect how the stressed material volume and structural capacity evolve as damage develops. We address this limitation by combining Hertzian elastomechanics, which govern the load- and curvature-dependent behavior of the subsurface stress field, with an Equivalent Damaged Volume (EDV) formulation that links geometry, stress, and microstructural evolution. In this framework, fatigue is governed by the growth of a representative damaged volume $V_d(t)$ under repetitive loading, making the degradation rate explicitly dependent on prior damage and on load-dependent weakening through a time-varying effective load rating $C_{\mathrm{eff}}(t)$, thereby capturing the progressive reduction in bearing capacity preceding macroscopic crack formation.\\
The Hertzian peak pressure for a ball contact is given by
\begin{equation}
\sigma_H = \sqrt[3]{\frac{6P (E^*)^2}{\pi^3 (R^*)^2}},
\end{equation}
which sets characteristic subsurface stress magnitudes. The baseline L--P damage intensity in seconds (cycles$^{-1}$) under current capacity $C_{\text{eff}}(t)$ is
\begin{equation}
k_f(t) = \left(\frac{P}{C_{\mathrm{eff}}(t)}\right)^{p} \frac{n}{60\times10^{6}},
\end{equation}
with $p\approx3$ for balls (or empirical alternative for rollers). To account for prior damage, we add an EDV term. For lumped modeling, the EDV growth rate is approximated as
\begin{equation}
\frac{dV_d}{dt} \approx \phi\left(\frac{P}{C_{\mathrm{eff}}(t)}\right)^{q} n,
\end{equation}
with $q>p$ reflecting the stronger sensitivity of damaged volume to contact stress peaks; $\phi$ is calibrated. The combined instantaneous fatigue damage rate is\\
\begin{equation}
\frac{dD_F}{dt} = k_f(t) + \beta\frac{dV_d}{dt} = \left(\frac{P}{C_{\text{eff}}(t)}\right)^{p} \frac{n}{60\times10^{6}} + \beta\phi\left(\frac{P}{C_{\text{eff}}(t)}\right)^{q} n.
\end{equation}
This form retains the empirical foundation of L--P while introducing a physically motivated, damage-accelerating EDV correction. The EDV term is small early in life ($\beta\phi\ll1$) and can dominate as $C_{\mathrm{eff}}(t)$ declines.

\subsubsection{Wear and Surface Roughness: Extended Archard with feedback}
Wear originates from sliding or rolling contact between surface asperities and is classically described by Archard’s law~\cite{kauzlarich2001archard}, which relates material loss to the applied load, sliding distance, and hardness but assumes fixed surface conditions. To capture the feedback mechanism observed in practice, we extend this framework by introducing surface roughness $R(t)$ as an internal state variable and by accounting for debris-induced abrasion. As the roughness increases, the local contact pressure peaks intensify. Brittle microasperities fracture more readily, accelerating wear. Moreover, the generated debris acts as a third-body abrasive cycle in which wear drives roughening, and roughening accelerates wear, captured through a time-dependent wear modification factor $W_{\mathrm{mod}}(t)$ that reduces the $C_{\mathrm{eff}}(t)$ as the surface progressively departs from its original geometry.\\
The wear volume rate is given by
\begin{equation}
\frac{dD_W}{dt} = \frac{A_v P}{H_{hard}}\frac{ds}{dt} + A_a R(t),
\end{equation}
where $A_v$ is the Archard wear coefficient, $H_{\mathrm{hard}}$ is the material hardness, $\mathrm{d}s/\mathrm{d}t$ is the sliding speed, and $A_a$ scales roughness-driven abrasive loss. Surface roughness evolves according to\\
\begin{equation}
\frac{dR}{dt} = \gamma_r \frac{dD_W}{dt} + \delta_c C_{debris}(t),
\end{equation}
where $\gamma_r$ links volume loss to roughness growth and $\delta_c$ captures contamination-induced roughening due to debris particles. A geometric degradation modifier is introduced to reduce the effective load-carrying capacity as the surface departs from its nominal geometry,
\begin{equation}
W_{mod}(t) = \bigl(1 + \eta V(t) + \zeta R(t)^2\bigr)^{-1}.
\end{equation}
This modifier is bounded such that $0 < W_{\mathrm{mod}} \leq 1$, penalizing both bulk material loss and amplified asperity-level stress.

\subsubsection{Lubricant Oxidation and Thermal Feedback}
Lubricant degradation in rolling element bearings is governed by thermally activated oxidative reactions that breakdown long-chain molecules, leading to a progressive loss of viscosity that follows Arrhenius kinetics and accelerates at elevated temperatures~\cite{peleg2012arrhenius}. We model this process by coupling chemical depletion, thermal effects, and the tribological response. Oxidation consumes a finite chemical reservoir $O_{\max}$, reducing viscosity and thinning the lubricant film that separates the rolling elements from the raceways. The resulting viscosity loss increases frictional dissipation, which increases the operating temperature and further accelerates oxidation, resulting in positive thermal feedback that can trigger runaway degradation under high load or insufficient cooling. By embedding these interactions within the thermal balance equation, the formulation captures how lubrication failure decreases the load-carrying capacity and amplifies fatigue and wear processes.\\
The oxidation state is modeled using first-order Arrhenius kinetics, approaching a saturation level $O_{\max}$,
\begin{equation}
\frac{dO}{dt} = k_o\bigl(O_{max}-O\bigr) \cdot e^{\Bigl(-\frac{E_a}{k_B T(t)}\Bigr)}.
\end{equation}
Kinematic viscosity decays with oxidation according to an empirical
\begin{equation}
    \nu(t) = \nu_0 \cdot e^{\left( -\alpha O(t) - \frac{E_{\mathrm{vis}}}{k_B T(t)} + \frac{E_{\mathrm{vis}}}{k_B T_0} \right)}
\end{equation}
Here, Evis is the activation energy for viscous flow (typically 20-40 kJ/mol for mineral oils, calibrated from lubricant datasheets showing viscosity-temperature curves), and T0 is a reference temperature (e.g., 313 K or 40°C). A lumped balance model tof hermal dynamics:
\begin{equation}
\frac{dD_O}{dt}=\frac{1}{m c_p}\Bigl[\,\mu_f P \omega - h A\bigl(T-T_a\bigr)+\xi\frac{dO}{dt}\,\Bigr],
\end{equation}
where $m c_p$ is the effective thermal mass, $\mu_f$ is the friction coefficient, which may be modeled as a function of viscosity in refined formulations, $hA$ is the effective heat transfer coefficient, and $\eta$ represents the exothermic contribution of oxidation reactions. The coupling is two-way: temperature $T$ controls the oxidation rate, oxidation reduces viscosity, and reduced viscosity increases frictional heating.

\subsubsection{Stochastic Micro-Contamination}
Contamination strongly accelerates bearing degradation, yet its evolution is inherently random due to particles originating from external ingress, internally generated wear debris, and intermittent release events associated with surface fracture. These contamination bursts exhibit non-Gaussian intermittent behavior that cannot be represented by deterministic dynamics. We therefore model debris concentration using a stochastic differential equation driven by a Wiener process, where the drift term captures deterministic generation from wear, and the diffusion term represents random fluctuations in particle generation, transport, agglomeration, and detachment. This stochastic formulation reflects variability arising from irregular operating conditions and lubricant flow instability, and captures the empirically observed burst-like increase in contamination that alters surface roughness, modifies contact mechanics, and rapidly accelerates fatigue progression.\\
Debris accumulation is modeled as a stochastic process to reflect bursty particle transport.
\begin{equation}
    dC_{debris}(t) = \rho \frac{dD_W}{dt} dt + \sigma_c d\mathcal{W}_t.
\end{equation}
where $\rho$ converts wear volume rate to particle concentration rate and $\sigma_c$ scales the noise intensity. This SDE produces sample paths with random spikes that match observed contamination bursts.

\subsubsection{Effective Load Rating and Mechanistic Coupling}
The effective load rating $C_{\mathrm{eff}}(t)$ provides a unifying state variable that couples the model’s four degradation pathways by allowing load-carrying capacity to evolve rather than remain static. Unlike classical rating, this formulation reflects how capacity degrades as lubrication deteriorates, surfaces wear, and contaminants accumulate. Lubricant breaks down the thin film, increasing contact stresses and accelerating fatigue, while wear-induced geometric distortion reduces contact conformity and amplifies local stress concentration. Contamination further intensifies abrasion and disrupts lubricant films, compounding both wear and fatigue. The combined effect is a dynamically shrinking $C_{\mathrm{eff}}(t)$ that drives nonlinear acceleration of damage near end-of-life, consistent with field observations: slow early degradation followed by rapid failure.\\
The effective load rating couples the four degradation processes:
\begin{equation}
C_{\mathrm{eff}}(t) = C_{load}\cdot L_{\text{life}}(t,T)\cdot W_{mod}(t)\cdot e^{\bigl(-\psi C_{debris}(t)\bigr)},
\end{equation}
where the lubrication life factor is
\begin{equation}
L_{\mathrm{\text{life}}}(t,T) = \frac{\nu(t)}{\nu_0}.
\end{equation}
Total degradation is
\begin{equation}
    \frac{dD_{\text{coupled}}}{dt} = \frac{dD_F}{dt} + \gamma_w \frac{dD_W}{dt} + \zeta_L \frac{dD_O}{dt}.
\end{equation}

\section{Experimental Details}

\begin{wrapfigure}{r}{0.6\textwidth}
\begin{minipage}{0.58\textwidth} 
\centering
\vspace{-4mm}
\captionsetup{type=table}
\caption{Data distributions of PRONOSTIA, XJTU-SY, and HUST datasets.}
\label{tab:data_dist}
\centering
\resizebox{0.99\textwidth}{!}{%
\begin{tabular}{ccccccc}
\hline
\textbf{Dataset} & \textbf{Condition} & \textbf{Frequency} & \textbf{Radial Load} & \textbf{Speed} & \textbf{\(\mathcal{X_{\text{IND}}}\)} &  \textbf{\(\mathcal{X_{\text{OOD/CV}}}\)}\\
\hline
\multirow{3}{*}{\textbf{PRONOSTIA}} & \textit{Condition 1}  & 100 Hz & 4 kN & 1800 rpm &  $4\sim7$ &  $1\sim3$ \\
& \textit{Condition 2}  & 100 Hz & 4.2 kN & 1650 rpm & \text{-} &  $1\sim7$ \\
& \textit{Condition 3}  & 100 Hz & 5 kN & 1500 rpm & \text{-} &  $1\sim3$ \\
\hline
\multirow{3}{*}{\textbf{XJTU-SY}} & \textit{Condition 1}  & 35 Hz & 12 kN & 2100 rpm & \text{-} &  $1\sim5$ \\
& \textit{Condition 2}  & 37.5 Hz & 11 kN & 2250 rpm & \text{-} &  $1\sim5$ \\
& \textit{Condition 3}  & 40 Hz & 10 kN & 2400 rpm & \text{-} &  $1\sim5$ \\
\hline
\multirow{3}{*}{\textbf{HUST}} & \textit{Condition 1}  & \text{-} & 0 W & \text{-} &   \text{-} & $1\sim5$ \\
& \textit{Condition 2}  & \text{-} & 200 W & \text{-} & \text{-} &  $1\sim5$ \\
& \textit{Condition 3}  & \text{-} & 400 W & \text{-} & \text{-} &  $1\sim5$ \\
\hline
\end{tabular}
}
\end{minipage}
\begin{minipage}{0.95\linewidth}
\footnotesize
\textbf{Note:} The PRONOSTIA dataset is utilized for training and generalization testing, while the XJTU-SY and HUST datasets are employed to evaluate cross-validation testing. (\text{-}) values are either not available or not utilized.
\end{minipage}
\vspace{-2mm}
\end{wrapfigure}

\subsection{Dataset Explanation}\label{appendix:description}
\textbf{PRONOSTIA:} dataset is a widely used benchmark in the field of condition monitoring and degradation estimation of REBs. Developed by Nectoux et al. \cite{nectoux2012pronostia} as part of the PRONOSTIA experimental platform, the dataset comprises 17 complete run‒to‒failure experiments conducted under accelerated wear conditions across three distinct operating regimes: 1800~rpm with a 4~kN radial load, 1650~rpm with a 4.2~kN load, and 1500~rpm with a 5~kN load, all recorded at a frequency of 100~Hz. Vibration data were captured via accelerometers mounted along both the horizontal and vertical axes and sampled at 25.6~kHz. Additionally, temperature measurements were recorded at a rate of 10 Hz. To train the models, we utilized data from four bearings (\(4\sim7 \in \mathcal{X}_{\text{IND}})\) operating under 100Hz4kN, incorporating both vibration and temperature data. To evaluate the model's generalization performance on \(\mathcal{X}_{\text{OOD}}\), the remaining bearings from all operating conditions were utilized. The summary of the dataset characteristics and distribution is provided in Table~\ref{tab:data_dist}.

\textbf{XJTU-SY:} dataset is a publicly available benchmark for degradation estimation of REBs. It was developed by Xi'an Jiaotong University in collaboration with Changxing Sumyoung Technology Company. The dataset consists of 15 complete run-to-failure experiments conducted under three operating conditions, defined by different combinations of rotational speed and radial load: 2100~rpm with a 12~kN load, 2250~rpm with an 11~kN load, and 2400~rpm with a 10~kN load. Vibration signals were collected using accelerometers mounted in both horizontal and vertical directions and sampled at 25.6~kHz. This dataset is used only for cross-validation results.

\textbf{HUST: }dataset was developed by Huazhong University of Science and Technology using a dedicated accelerated life test platform. The dataset comprises five bearings tested under three different load conditions: 0~W, 200~W, and 400~W. Vibration signals were acquired using accelerometers mounted along the horizontal and vertical directions and sampled at 25.6~kHz. Data were recorded continuously throughout the bearing lifecycle until failure, providing complete degradation trajectories. This dataset is used only for cross-validation experiments.

\begin{wrapfigure}{r}{0.6\textwidth}
\vspace{-4mm}
\begin{minipage}{0.58\textwidth}
\begin{algorithm}[H]
\caption{Time‒frequency representation extraction algorithm}\label{algo:tfr}
\footnotesize
\begin{algorithmic}
\Require windowed signal $I_{w}$, critical frequency $f_c$, operating frequency $f_o$, sampling period $T_{sampling}$, windowed physical constraints ($t_w,T_w$)
\State $a_{min} = \frac{f_c}{f_{max}\cdot T_{sampling}}, \quad a_{{max}} = \frac{f_c}{f_{\text{min}}\cdot T_{sampling}}$,
\State $a_{scale} \in [a_{min}, a_{max}]$
\State $I_{{TFR}} \gets \{\}$

\For{($i_w ,t_n ,T_n$) in ($I_{w},t_w ,T_w$):}
        \State Compute wavelets as: \(\Gamma_{iw}(a,b) = \int_{-\infty}^{\infty} i_w \psi^* \left(\frac{t - b}{a} \right) dt.\)
        \State Compute Energy as: \(E = \sum_{m=1}^{M} \left| \Gamma_{iw}(a,b) \right|^2\)
        \State Compute Dominant frequency as: \(f_{d} = a_{\text{scale}} \left[ \arg\max(E) \right]\).
        \State Compute Entropy as: \(h = -\sum_{i=m}^{M} P(i_w(t)) \log P(i_w(t))\).
        \State Compute Kurtosis as: \(K = \frac{\mathbb{E}[(i_w(t) - \mu)^4]}{\sigma^4}\).
        \State Compute Skewness as: \(s_k = \frac{\mathbb{E}[(i_w(t) - \mu)^3]}{ \sigma^3}\).
        \State Compute mean as: \(\mu = \frac{1}{M} \sum_{m=1}^{M} i_w(m)\).
        \State Compute standard deviation as: \(\sigma = \sqrt{\frac{1}{M} \sum_{i=1}^{M} \left(i_w(m) - \mu \right)^2}\).
        \State $X_n \gets [\log(E), f_{d}, h, K, sk, \mu, \sigma] $
\EndFor
\State \Return $I_{TFR} = Concat(X_{1}, X_{2} \ldots X_{N_s},t_n,T_n)$
\end{algorithmic}
\end{algorithm}
\end{minipage}
\vspace{-5mm}
\end{wrapfigure}

\subsection{Data Curation}\label{appendix:curation}
The condition monitoring data of rotating machinery typically comprises 1-D nonstationary vibration signals acquired from multiple sensors. To facilitate time‒frequency analysis, the raw signals \( x(t) \) are first segmented using a time-based rectangular windowing function \( w(t) \), allowing improved localization of transient features across the time and frequency domains. Each segmented signal \( x_w(t) = x(t) \cdot w(t) \) is then subjected to a continuous wavelet transform (CWT) \cite{aguiar2014continuous} using the Morlet wavelet \cite{bussow2007algorithm} \( \psi(t) \) as the mother wavelet. The Morlet wavelet is selected because it resembles the impulse response of localized bearing faults \cite{tang2010wind,zhu2018estimation}, which enhances sensitivity to fault-induced transients \cite{razzaq2025carle}. The CWT is defined as:
\begin{equation}
W(a, b) = \int_{-\infty}^{\infty} x_w(t) \frac{1}{\sqrt{a}} \psi^*\left( \frac{t - b}{a} \right) \, dt,
\end{equation}
where \( a \) and \( b \) denote the scale and translation parameters, respectively. From the resulting time‒frequency representation (TFR), a set of statistical and domain-relevant features is extracted to characterize the bearing’s operational condition. These features are computed in both the time and frequency domains and are selected to maintain physical interpretability while maximizing discriminative power. A detailed summary of all the extracted features, their mathematical formulations, and their physical significance is provided in Table~\ref{tab:feat}. Algorithm \ref{algo:tfr} provides the full method for extracting the TFR features.

\begin{table}[ht]
\centering
\caption{Summary of time‒frequency features, formulas, domains, and physical meanings}
\vspace{1mm}
\resizebox{0.85\textwidth}{!}{%
\begin{tabular}{cccp{6cm}}  
\midrule
\textbf{Feature}              & \textbf{Formula}                                                       & \textbf{Domain}        & \textbf{Physical Meaning}                                         \\
\midrule
\textbf{Energy (\(E\))}        & \( E = \sum_{m=1}^{M} \left| \Gamma_{i}(a,b) \right|^2 \)           & Frequency              & Measures vibrational activity. Increases suggest wear or defects (e.g., spalling).                                         \\ \midrule
\textbf{Dominant Frequency (\(f_d\))} & \( f_d = a_{\text{scale}}(\text{argmax}(E)) \)                        & Frequency         & Identifies frequency with the highest energy, which is useful for detecting faults such as bearing cracks.\\ \midrule
\textbf{Entropy (\(h\))}       & \( h = -\sum_{i=1}^{K} P(i) \log P(i) \)                           & Time              & Quantifies randomness in vibrations. Higher values indicate irregular defects or friction.                                     \\ \midrule
\textbf{Kurtosis (\(K\))}      & \( K = \frac{\mathbb{E}[(i - \mu)^4]}{\sigma^4} \)                   & Time              & Detects extreme signal spikes. High values indicate localized defects (e.g., cracks).                                           \\ \midrule
\textbf{Skewness (\(s_k\))}    & \( s_k = \frac{\mathbb{E}[(i - \mu)^3]}{\sigma^3} \)                 & Time              & Measures signal asymmetry. Positive skew suggests unidirectional impacts, negative skew suggests crack initiation.                      \\ \midrule
\textbf{Mean (\(\mu\))}        & \( \mu = \frac{1}{N} \sum_{m=1}^{M} i(m) \)                          & Time              & Baseline vibrational level. Increases suggest wear or faults (e.g., cage failure).                                       \\ \midrule
\textbf{Standard Deviation (\(\sigma\))} & \( \sigma = \sqrt{\frac{1}{M} \sum_{i=1}^{m} (i(m) - \mu)^2} \) & Time              & Measures signal variability. High values indicate instability or faults such as looseness or contamination.\\ \midrule
\end{tabular}}
\label{tab:feat}
\end{table}

\subsection{Evaluation Metrics}\label{appendix:metrics}
In this study, we first evaluate the prediction accuracy of the developed approaches using commonly employed metrics such as mean squared error (MSE), mean absolute error (MAE), and the Score~\cite{nectoux2012pronostia}. Existing metrics for evaluating uncertainty in deep learning models, including negative log-likelihood (NLL) and root mean square calibration error (RMSCE), primarily assess overall predictive uncertainty and the model’s confidence in its predictions. However, in safety-critical and risk-sensitive applications, such as degradation estimation, it is essential that the model’s predictive uncertainty effectively informs decision-makers about the reliability of predictions, particularly regarding potential failures. Therefore, we propose a distance-aware calibration performance metric.

\textbf{Distance-Aware-Coefficienct (DAC):} Consider a predictive distribution \( p(y|x) \) trained on \( \mathcal{X}_{\text{IND}}\), where \( (\|\cdot\|_x) \) denotes the input data manifold equipped with an appropriate metric. We define \( p(y|x) \) as input distance-aware if there exists a summary statistic \( a(x) \) of the predictive distribution that quantifies uncertainty in a manner that reflects the distance between the test input \( x \) and the training data manifold:
\begin{equation}
a(x) = g_m(d(x, \mathcal{X}_{\text{IND}})),
\end{equation}

where \( g_m \) is a monotonic function and \(d(x, \mathcal{X}_{\text{IND}}) = \mathbb{E}_{x' \sim \mathcal{X}_{\text{IND}}}[\|x - x'\|_x]\) represents the expected distance from \( x \) to \(\mathcal{X}_{\text{IND}}\) calculated using Euclidean distance. To measure the sensitivity of the quantified uncertainty with respect to this distance, we introduce the distance-aware coefficient (DAC), defined as the Pearson correlation coefficient \cite{cohen2009pearson} between the distance \(d(x) \) and the predictive uncertainty \( \sigma \). Formally, the DAC is given by:
\begin{equation}
\text{DAC} = \frac{\sum_{i=1}^{N} (d_i - \bar{d})(\sigma_i - \bar{\sigma})}{\sqrt{\sum_{i=1}^{N} (d_i - \bar{d})^2} \sqrt{\sum_{i=1}^{N} (\sigma_i - \bar{\sigma})^2}},
\end{equation}
where \( d_i \) is the distance of the \( i \)-th test sample to the training dataset, \( \sigma_i \) is its associated predictive uncertainty, and \( \bar{d} \), \( \bar{\sigma} \) are their respective sample means. Ideally, test points that are farther from the training dataset should exhibit higher predictive uncertainty, and vice versa. A higher DAC value indicates superior performance of the uncertainty quantification mechanism in capturing input-dependent reliability.

\begin{table}[htbp]
    \centering
    \caption{Physical parameters and values  dataset}
    \resizebox{0.78\textwidth}{!}{%
    \begin{tabular}{@{}l|c|c@{}}
    \toprule
    \textbf{Physical variable} & \textbf{PRONOSTIA} & \textbf{XJTU-SY} \\ \midrule

    $C$ (Dynamic load rating) & 4000 N & 12000 N \\
    $p$ (Fatigue law exponent) & 3.0 & 3.0 \\
    $q$ (EDV load exponent) & 4.0 & 4.0 \\
    $\beta$ (Fatigue--EDV coupling coefficient) & $1 \times 10^{-6}$ & $1 \times 10^{-6}$ \\
    $\phi$ (EDV growth coefficient) & $1 \times 10^{-5}$ & $1 \times 10^{-5}$ \\

    $D_m$ (Ball diameter) & 0.025 m & 0.035 m \\
    $k_B$ (Boltzmann constant) & $8.617 \times 10^{-5}$ eV/K & $8.617 \times 10^{-5}$ eV/K \\
    $E_a$ (Activation energy) & 0.1 eV & 0.1 eV \\
    $E_{\text{vis}}$ (Viscosity activation energy) & 0.1 eV & 0.1 eV \\
    $T_0$ (Reference temperature) & 298 K & 298 K \\
    $T_a$ (Ambient temperature) & 298 K & 298 K \\

    $\alpha$ (Viscosity degradation coefficient) & $1 \times 10^{-5}$ & $1 \times 10^{-5}$ \\
    $\nu_0$ (Baseline lubricant viscosity) & $1 \times 10^{-5}$ & $1 \times 10^{-5}$ \\
    $k_o$ (Oxidation rate coefficient) & $1 \times 10^{-4}$ & $1 \times 10^{-4}$ \\
    $O_{\max}$ (Maximum oxidation level) & 1.0 & 1.0 \\

    $A_v$ (Archard wear coefficient) & $1 \times 10^{-5}$ & $1 \times 10^{-6}$ \\
    $A_a$ (Roughness-induced wear coefficient) & $1 \times 10^{-6}$ & $1 \times 10^{-6}$ \\
    $H_{\text{hard}}$ (Material hardness) & $1.5 \times 10^{9}$ Pa & $1.5 \times 10^{9}$ Pa \\
    $\gamma_r$ (Roughness evolution from wear) & $1 \times 10^{-3}$ & $1 \times 10^{-3}$ \\
    $\delta_c$ (Roughness evolution from debris) & $1 \times 10^{-5}$ & $1 \times 10^{-5}$ \\
    $\rho$ (Debris generation coefficient) & $1 \times 10^{6}$ & $1 \times 10^{6}$ \\

    $\eta$ (Volume saturation coefficient) & $1 \times 10^{6}$ & $1 \times 10^{6}$ \\
    $\zeta$ (Roughness saturation coefficient) & $1 \times 10^{12}$ & $1 \times 10^{12}$ \\

    $\gamma_w$ (Wear contribution weight) & 0.5 & 0.1 \\
    $\zeta_L$ (Thermal contribution weight) & 0.5 & 0.1 \\
    $m c_p$ (Mass times specific heat) & $3.77 \times 10^{6}$ & $3.77 \times 10^{6}$ \\
    $\mu_f$ (Friction coefficient) & 0.005 & 0.005 \\
    $h_A$ (Heat transfer coefficient) & 5.0 & 5.0 \\
    $\xi$ (Oxidation exothermic contribution) & 100.0 & 100.0 \\

    \bottomrule
    \end{tabular}}
    \label{tab:physical}
\end{table}

\end{document}